\documentclass[conference]{IEEEtran}

\usepackage{times}
\usepackage{epsfig}
\usepackage{graphicx}
\usepackage{amsmath}
\usepackage{amssymb}
\usepackage{multirow}
\usepackage{setspace}
\usepackage{subfigure}
\usepackage{balance}
\usepackage{url}

\usepackage{array}
\newcolumntype{V}{>{$\vcenter\bgroup\hbox\bgroup}c<{\egroup\egroup$}}
\def\Hline{\noalign{\hrule height 4\arrayrulewidth}}

\graphicspath{{./imgs/}}

\begin{document}

\title{The Adaptable Buffer Algorithm for High Quantile Estimation in Non-Stationary Data Streams}

\author{
    Ognjen Arandjelovi\'c$^\dag$ , Duc-Son Pham$^\ddag$ , and Svetha Venkatesh$^\dag$ \\~\\
    \begin{tabular}{cc}
      $^\dag$ Centre for Pattern Recognition \& Data Analytics & $^\ddag$ Department of Computing\\
      ~Deakin University                                     & ~~Curtin University\\
      ~Geelong VIC 3216                                      & ~~Perth WA 6845\\
      ~Australia                                             & ~~Australia\\~\\
    \end{tabular}
}

\maketitle

\begin{abstract}
The need to estimate a particular quantile of a distribution is an important problem which frequently arises in many computer vision and signal processing applications. For example, our work was motivated by the requirements of many semi-automatic surveillance analytics systems which detect abnormalities in close-circuit television (CCTV) footage using statistical models of low-level motion features. In this paper we specifically address the problem of estimating the running quantile of a data stream with non-stationary stochasticity when the memory for storing observations is limited. We make several major contributions: (i) we derive an important theoretical result which shows that the change in the quantile of a stream is constrained regardless of the stochastic properties of data, (ii) we describe a set of high-level design goals for an effective estimation algorithm that emerge as a consequence of our theoretical findings, (iii) we introduce a novel algorithm which implements the aforementioned design goals by retaining a sample of data values in a manner adaptive to changes in the distribution of data and progressively narrowing down its focus in the periods of quasi-stationary stochasticity, and (iv) we present a comprehensive evaluation of the proposed algorithm and compare it with the existing methods in the literature on both synthetic data sets and three large `real-world' streams acquired in the course of operation of an existing commercial surveillance system. Our findings convincingly demonstrate that the proposed method is highly successful and vastly outperforms the existing alternatives, especially when the target quantile is high valued and the available buffer capacity severely limited.
\end{abstract}

\IEEEpeerreviewmaketitle

\section{Introduction}
Quantile estimation is of pervasive importance across a variety of signal processing applications. It is used extensively in data mining, simulation modelling~\cite{JainChla1985}, database maintenance, risk management in finance~\cite{AdleFeldTaqq1998,SgouYaoYast2013}, and the analysis of computer network latencies~\cite{BuraSuri2009,CormJohnKornMuth+2004}, amongst others. A particularly challenging form of the quantile estimation problem arises when the desired quantile is high-valued (close to unity) and when data needs to be processed as a stream, with limited memory capacity. An illustrative practical example of when this is the case is encountered in CCTV-based surveillance systems~\cite{AranPhamVenk2015b}. In summary, as various types of low-level observations related to events in the scene of interest arrive in real-time, quantiles of the corresponding statistics for time windows of different durations are needed in order to distinguish `normal' (common) events from those which are in some sense unusual and thus require human attention. The amount of incoming data is extraordinarily large and the capabilities of the available hardware highly limited both in terms of storage capacity and processing power.

\subsection{Previous work}\label{ss:prev}
Unsurprisingly, the problem of estimating a quantile of a set has received considerable attention, much of it in the realm of theoretical research. In particular, a substantial amount of work has focused on the study of asymptotic computational complexity of quantile estimation algorithms~\cite{GuhaMcGr2009,MunrPate1980}. An important result emerging from this corpus of work is the proof by Munro and Paterson~\cite{MunrPate1980} that the working memory requirement of any algorithm that determines the median of a set by making at most $p$ sequential passes through the input is $\Omega(n^{1/p})$ (i.e.\ asymptotically growing at least as fast as $n^{1/p}$). This implies that the exact computation of a quantile requires $\Omega(n)$ working memory. Therefore a single-pass algorithm, required to process streaming data, will necessarily produce an estimate and not be able to guarantee the exactness of its result.

Most of the quantile estimation algorithms developed for use in practice are not single-pass algorithms and thus cannot be applied to streaming data~\cite{GuraSriv1990}. On the other hand, many single-pass approaches focus on the exact computation of the quantile and therefore, as explained previously, demand the $O(n)$ storage space which is clearly an unfeasible proposition in the context we consider in the present paper; this includes the work by Greenwald and Khanna~\cite{GreeKhan2001} who described an $O(n)$ method efficient in the sense that it attains the asymptotic minimum in the space requirement as a function of the permissible error in the desired quantile estimate. Amongst the few methods described in the literature which satisfy the practical constraints of interest in the present paper are the histogram-based method of Schmeiser and Deutsch~\cite{SchmDeut1977} (with a similar approach described by McDermott \textit{et al.}~\cite{McDeBabuLiecLin2007}), and the $P^2$ algorithm of Jain and Chlamtac~\cite{JainChla1985}. Schmeiser and Deutsch maintain a preset number of bins, scaling their boundaries to cover the entire data range as needed and keeping them equidistant. Jain and Chlamtac attempt to maintain a small set of \textit{ad hoc} selected key points of the data distribution, updating their values using quadratic interpolation as new data arrives. Various random sample methods, such as that described by Vitter~\cite{Vitt1985}, and Cormode and Muthukrishnan~\cite{CormMuth2005}, use different sampling strategies to fill the available buffer with random data points from the stream, and estimate the quantile using the distribution of values in the buffer. Lastly, the recently proposed algorithm of Arandjelovi\'c \textit{et al.}~\cite{AranPhamVenk2014} employs an adaptable quasi-maximum entropy histogram; their approach is discussed further in Section~\ref{ss:nonstoch}.

In addition to the \textit{ad hoc} elements of the existing algorithms for quantile estimation on streaming data, which itself is a sufficient cause for concern when the algorithms need to be deployed in applications which demand high robustness and well understood failure modes, it is also important to recognize that an implicit assumption underlying these approaches (with the exception of the algorithm of Arandjelovi\'c \textit{et al.}~\cite{AranPhamVenk2014}; see Section~\ref{ss:nonstoch}) is that the data is governed by a stationary stochastic process. The assumption is often invalidated in real-world applications.

\section{Proposed algorithm}
We begin by a formalization of the notion of a quantile, follow by a derivation of the key results underlying our contribution, and finally a describe the proposed algorithm.

\subsection{Quantiles}
Let $p$ be the probability density function of a real-valued random variable $X$. Then the $q$-quantile $v_q$ of $p$ is defined as~\cite{KennKeep1962}:
\begin{align}
  \int_{-\infty}^{v_q} p(x)~dx = q.
\end{align}
Similarly, the $q$-quantile of a finite set $D$ can be defined as:
\begin{align}
  \left|\{ x~:~ x \in D \text{ and  } x \leq v_q\}\right| \leq q \times |D|.
  \label{e:qSet}
\end{align}
In other words, the $q$-quantile is the smallest value below which $q$ fraction of the total values in a set lie. The concept of a quantile is thus intimately related to the tail behaviour of a distribution.

\subsection{Challenges of non-stochasticity}\label{ss:nonstoch}
\begin{figure}[htb]
  \centering
  \includegraphics[width=0.48\textwidth]{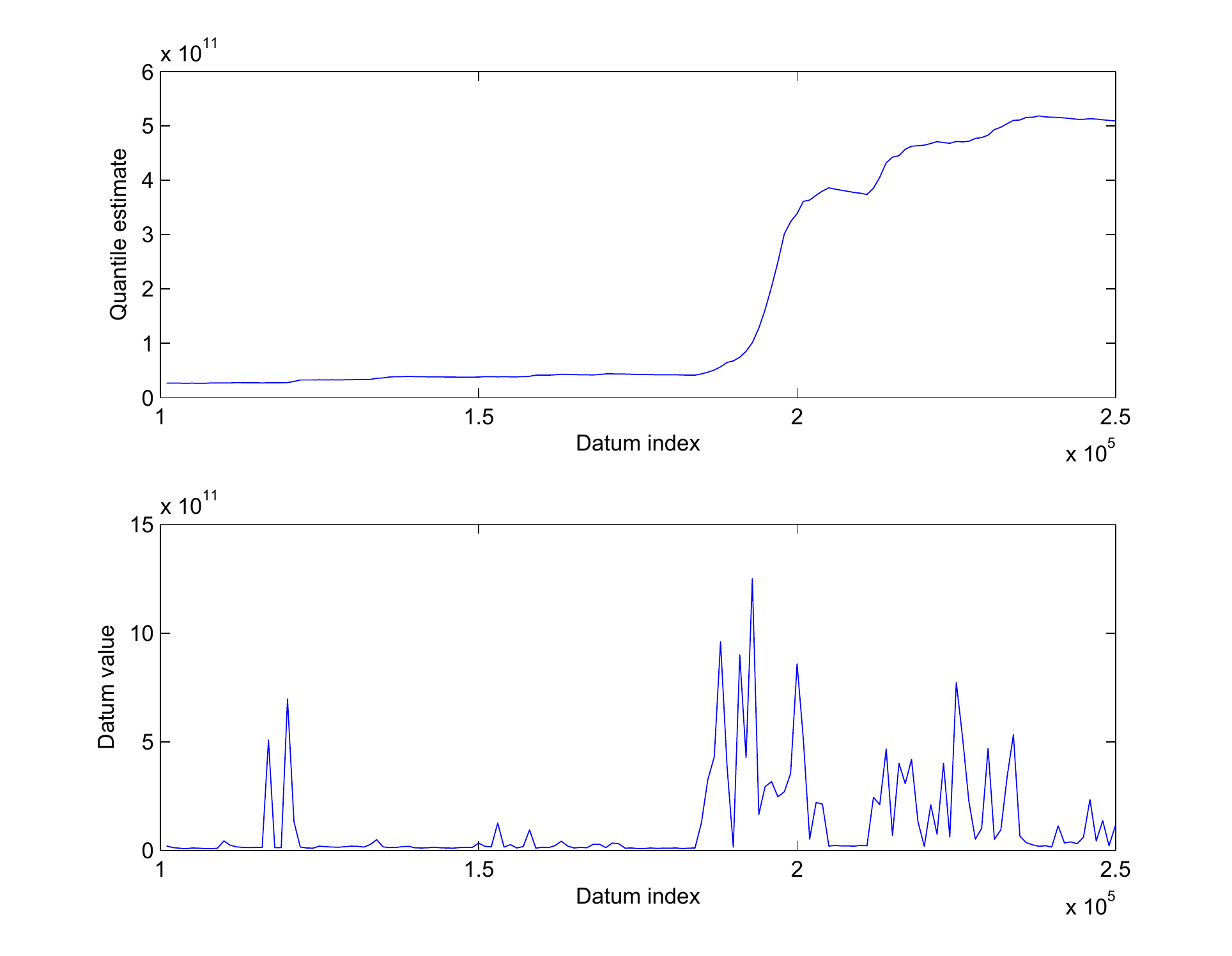}
  \caption{An example of a rapid change in the value of a quantile (specifically the 0.95-quantile in this case) on a real-world data stream used for surveillance video analysis (see Section~\ref{sss:dataReal}). }
  \label{f:change}
\end{figure}

In this work our aim is to develop a method for quantile estimation applicable not only to streams which exhibit stationary stochasticity but also to the all-encompassing set of streams which includes those with non-stationary data. It is a straightforward consequence of potential non-stationarity that at no point in time can it be assumed that the historical distribution of data values is representative of the future distribution of the stream data. This is true regardless of how much historical data has been seen. Thus, the value of a particular quantile can change greatly and rapidly, in either direction (i.e.\ increase or decrease). This is illustrated on an example, extracted from a real-world data set used for surveillance video analysis (the full data corpus is used for comprehensive evaluation of different methods in Section~\ref{s:eval}), in Figure~\ref{f:change}. In particular, the top plot in this figure shows the variation of the ground-truth 0.95-quantile which corresponds to the data stream shown in the bottom plot. Notice that the quantile exhibits little variation over the course of approximately the first 75\% of the duration of the time window (the first 190,000 data points). This corresponds to a period of little activity in the video from which the data is extracted (see Section~\ref{sss:dataReal} for a detailed explanation). Then, the value of the quantile increases rapidly for over an order of magnitude -- this is caused by a sudden burst of activity in the surveillance video and the corresponding change in the statistical behaviour of the data.

It may appear to be the case that to be able to adapt to such unpredictable variability in input it is necessary to maintain an approximation of the entire distribution of historical data. Indeed, this is argued in a recent work which introduced the Data-Aligned Maximum Entropy Histogram algorithm for quantile estimation from streams~\cite{AranPhamVenk2014}. The method employs a histogram of a fixed length, determined by the available working memory, which adjusts bin boundary values in a manner which maximizes the entropy of the corresponding estimate of the historical data distribution.

Although it is true that the change in the value of a specific quantile may be of an arbitrary large magnitude, in this paper we show that its specific value in a particular stream is nevertheless constrained. Succinctly put, this is a consequence of the fact that although the stream data may be considered as being drawn from a continuous probability density function (which may change with time) the information available to our algorithm inherently comprises discrete quanta: individual data points.

\subsection{Constraints: key theoretical results}\label{sss:constraints}
Consider a stream of values $x_1, x_2, \ldots, x_n$. For the time being let us assume that there are no repeated values in the stream i.e.\ $\forall i,j.~x_i=x_j \Longrightarrow i=j$. Then there is an indexing function $f(\ldots)$ such that $x_{f_n(1)} < x_{f_n(2)} < \ldots < x_{f_n(n)}$. Let $x_{q(n)}=x_{f(k)}$ be the current estimate of a particular quantile $q$ of interest. Consider $\Delta k$, the change in $k$ that the arrival of a new datum $x_{n+1}$ effects. By definition given in Equation~\ref{e:qSet}:
\begin{align}
     \Delta k = &\lfloor (1-q) \times (n+1) \rfloor - \lfloor (1-q) \times n \rfloor.
\end{align}
Exploiting simple properties of the flood function then leads to the following series of inequalities and an upper bound on the value of $\Delta k$:
\begin{align}
     \Delta k = &\lfloor (1-q) \times (n+1) \rfloor - \lfloor (1-q) \times n \rfloor \\
           \leq &(1-q) \times (n+1)- \lfloor (1-q) \times n \rfloor \\
              = &(1-q) \times n- \lfloor (1-q) \times n \rfloor + (1-q) \\
              < &1 + (1-q) = 2 - q < 2,
\end{align}
Since $\Delta k$ has to be an integer:
\begin{align}
     \Delta k   & \leq 1.
     \label{e:constraint1}
\end{align}
A similar sequence of steps can also give us the lower bound on $\Delta k$:
\begin{align}
     \Delta k = & \lfloor (1-q) \times (n+1) \rfloor - \lfloor (1-q) \times n \rfloor \\
           \geq & \lfloor (1-q) \times (n+1) \rfloor - (1-q) \times n \\
              = & \lfloor (1-q) \times (n+1) \rfloor - (1-q) \times (n+1) + (1-q)\\
           \geq & -1 + (1-q) = -q,\\
     \text{and since } &\Delta k \text{ has to be an integer:} \notag \\
     \Delta k   & \geq -q \geq 0.
     \label{e:constraint2}
\end{align}
Finally, combining the two results gives:
\begin{align}
   0 \leq \Delta k \leq 1.
   \label{e:constraint}
\end{align}
Thus, rather remarkably at first sight, regardless of the value of the new datum $x_{n+1}$, the change in the index in the sorted stream that references the correct quantile value can either remain unchanged or increase by one. This shows that while the observation made in Section~\ref{ss:nonstoch}, that the \emph{value} of the quantile estimate may exhibit an arbitrarily large change, it is nonetheless constrained to the specific values of the stream just below or just above the previous (current) estimate -- a consequence of the inherently quantized nature of data which comprises the stream (discrete data points). This insight motivates to propose the following three key ideas for an effective and efficient algorithm:
\begin{itemize}
  \item the buffer should store a list of monotonically increasing stream values
  \item the current quantile should be as close to the centre of the buffer as possible
  \item the spread of buffer values should decrease when the estimate is unchanging
\end{itemize}
Lastly, we we will show in the next section, the assumption of non-repeating data can be removed by employing a representation which does not store repeated observations but nevertheless keeps track of repetition using an auxiliary data structure.

\subsection{Targeted Adaptable Sample algorithm}\label{sss:alg}
Having laid out the key theoretical results underpinning our approach we are now in the position to introduce our quantile estimation algorithm. At the heart of the proposed method is a data structure which comprises two parts. The first of these is an ordered list of data points $b_1 < b_2 < \ldots < b_m$, where $m$ is the buffer capacity (size), selected from the input data stream. The second part of the structure is auxiliary information, a sequence $a_1, a_2, \ldots, a_m$ of values, associated with the selected data. Specifically, for each remembered datum $b_i$ we also maintain an estimate $a_i$ of the number of historical data points whose value is lower than that of the datum i.e.\ after the processing of $n$ data points, $a_i$ is the estimate of $| \{ x_j ~|~ x_j < b_i, j=1,\ldots,n \} |$.

The first $m$ unique data points are simply stored in the buffer in the increasing order; the associated auxiliary counts $a_1,\ldots,a_m$ can be computed exactly. With the arrival of each new datum $x_{n+1}$ thereafter, the following sequence of steps takes place. Firstly, if the value of the new datum is already present in the buffer, the auxiliary counts corresponding to greater buffer values are incremented by one. Otherwise, the index $k$ into the buffer of the current quantile is determined by finding the lowest element $b_k$ in the buffer such that $a_k / n \geq 1-q$, where $q$ is the target quantile. Then if the new datum is smaller than the current quantile estimate, i.e.\ $x_{n+1} < b_k$, and either $k<\lfloor m/2 \rfloor$ or $x_{n+1} > b_1$, the new datum $x_{n+1}$ is inserted into the buffer and the largest value in the buffer, $b_m$, discarded. The former case reinforces the central positioning of the current quantile estimate, while the latter acts so as to decrease the spread of values within the buffer. The auxiliary count corresponding to the newly inserted datum is initialized by linearly interpolating between the counts of buffer values between which the datum is inserted. Auxiliary counts corresponding to lower valued buffer elements are left unchanged while those corresponding to higher valued elements are increased by one. Similarly, if the new datum is greater than the current quantile estimate, i.e.\ $x_{n+1} > b_k$, and either $k>\lfloor m/2 \rfloor$ or $x_{n+1} < b_m$, the new datum $x_{n+1}$ is inserted into the buffer and the smallest value in the buffer, $b_1$, discarded.

\section{Evaluation}\label{s:eval}
We now turn our attention to the evaluation of the proposed algorithm. In particular, to assess its effectiveness and compare it with the algorithms described in the literature, in this section we report its performance on three large `real-world' data streams.

\subsection{Real-world surveillance data}\label{sss:dataReal}
\begin{figure*}[htb]
  \centering
  \subfigure[Data stream~1]{\includegraphics[width=0.42\textwidth]{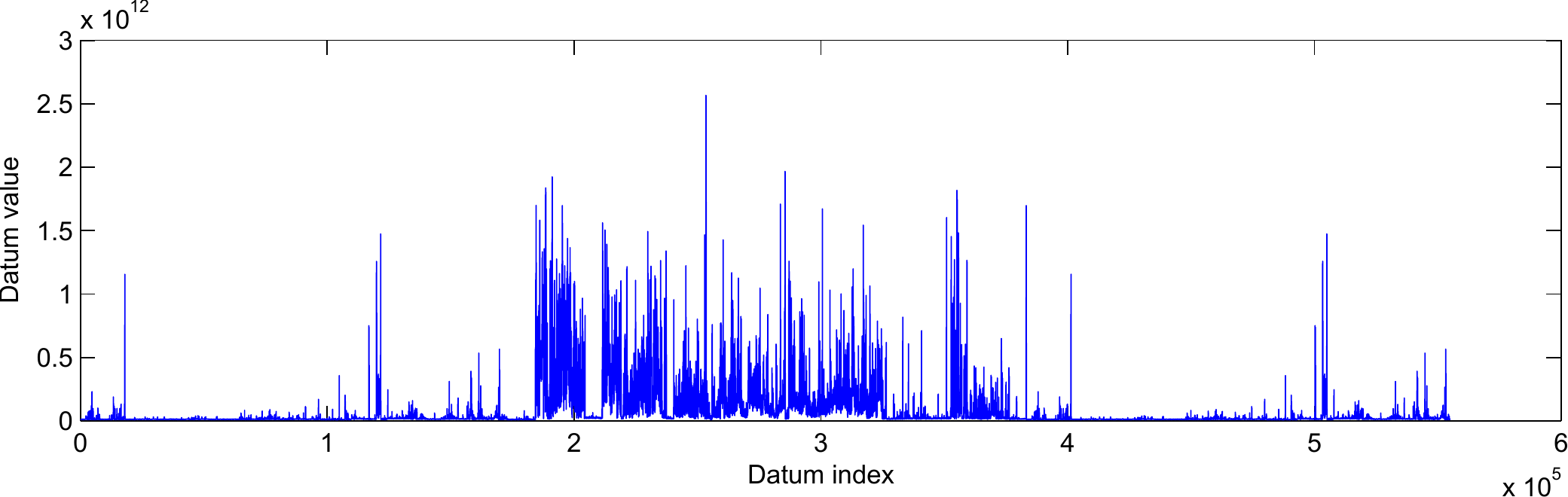}}
  \subfigure[Data stream~2]{\includegraphics[width=0.5\textwidth]{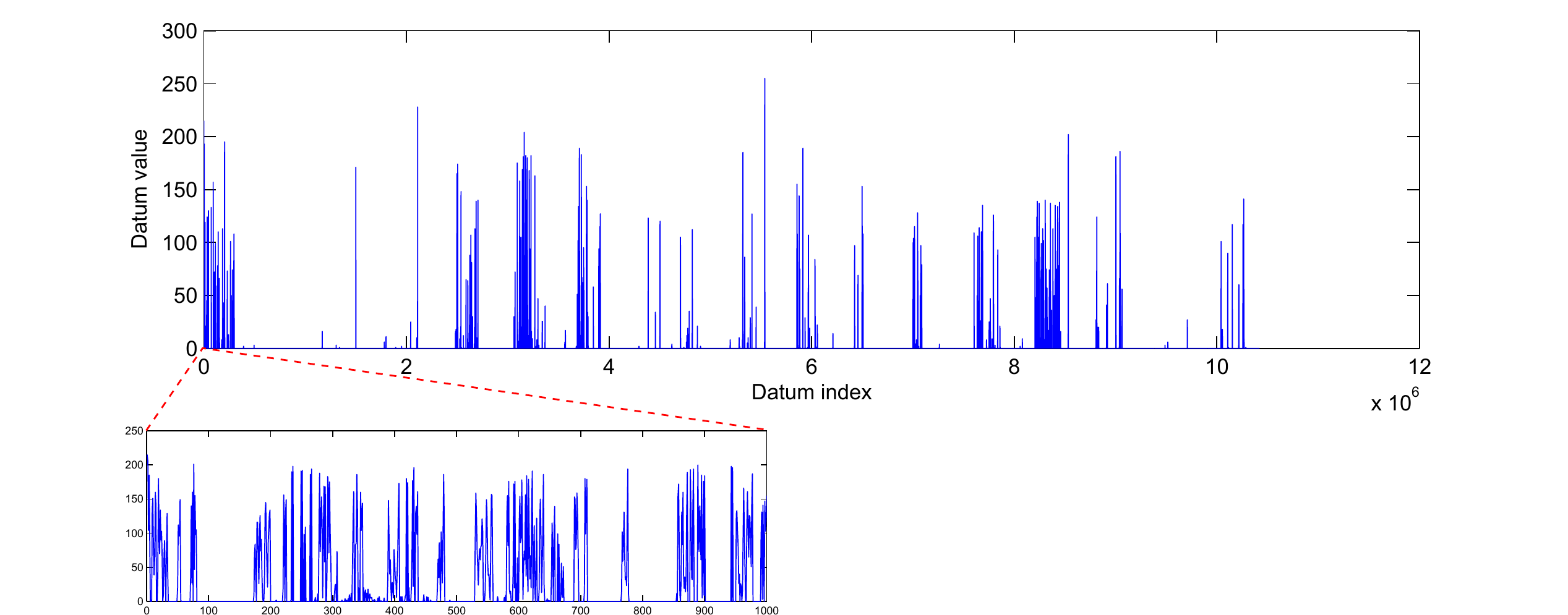}}
  \subfigure[Data stream~3]{\includegraphics[width=1\textwidth]{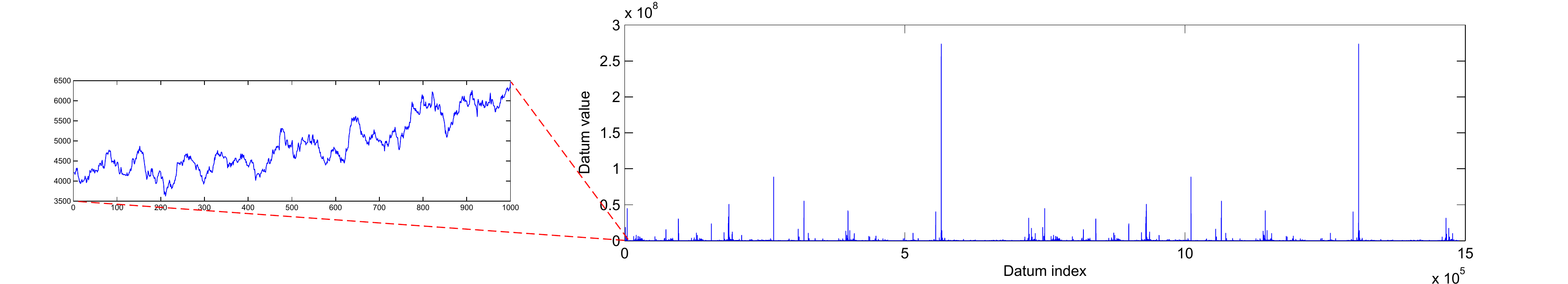}}
  \vspace{3pt}
  \caption{ The three large `real-world' data streams used in our evaluation.  }
  \label{f:streams}
  \vspace{3pt}
\end{figure*}

Computer-assisted video surveillance data analysis is of major commercial and law enforcement interest. On a broad scale, systems currently available on the market can be grouped into two categories in terms of their approach. The first group focuses on a relatively small, predefined and well understood subset of events or behaviours of interest such as the detection of unattended baggage, violent behaviour, etc~\cite{Phil,LaveKhanThur2007}. The narrow focus of these systems prohibits their applicability in less constrained environments in which a more general capability is required. These approaches tend to be computationally expensive and error prone, often requiring fine tuning by skilled technicians. This is not practical in many circumstances, for example when hundreds of cameras need to be deployed as often the case with CCTV systems operated by municipal authorities. The second group of systems approaches the problem of detecting suspicious events at a semantically lower level~\cite{iCet,PhamAranVenk2015,MartAran2010,Aran2011a,AranPhamVenk2015a}. Their central paradigm is that an unusual behaviour at a high semantic level will be associated with statistically unusual patterns (also `behaviour' in a sense) at a low semantic level -- the level of elementary image/video features. Thus methods of this group detect events of interest by learning the scope of normal variability of low-level patterns and alerting to anything that does not conform to this model of what is expected in a scene, without `understanding' or interpreting the nature of the event itself. These methods uniformly start with the same procedure for feature extraction. As video data is acquired, firstly a dense optical flow field is computed using the well-known method of Lucas and Kanade~\cite{LucaKana1981}. Then, to reduce the amount of data that needs to be processed, stored, or transmitted, a thresholding operation is performed. This results in a sparse optical flow field whereby only those flow vectors whose magnitude exceeds a certain value are retained; non-maximum suppression is applied here as well~\cite{Pham2010}. Normal variability within a scene and subsequent novelty detection are achieved using various statistics computed over this data. The data streams, shown partially in Figure~\ref{f:streams}, correspond to the values of such statistics (their exact meaning is proprietary and has not been made known fully to the authors of the present paper either). Observe the non-stationary nature of the streams which is evident both on the long and short time scales (magnifications are shown for additional clarity and insight). Table~\ref{t:data} provides a summary of some of the key features of the three data sets acquired in the described manner and used for the evaluation in this paper.

\begin{table}[htb]
  \vspace{7pt}
  \caption{Key statistics of the three real-world data sets used in our evaluation. }
  \renewcommand{\arraystretch}{1.55}
  \vspace{1pt}
  \centering
  \small
  \begin{tabular}{l|ccc}
  \Hline
  Data set  & Data points & Mean value & Standard deviation\\
  \hline
  Stream~1  &    $555,022$ & $7.81\times 10^{10}$ & $1.65\times 10^{11}$\\
  Stream~2  & $10,424,756$ & $2.25$               & $15.92$ \\
  Stream~3  &  $1,489,618$ & $1.51\times 10^{5}$  & $2.66\times 10^{6}$\\
  \Hline
  \end{tabular}
  \label{t:data}
\end{table}

\subsection{Results}
We compared the performances of our algorithm and the four alternatives from the literature described in Section~\ref{ss:prev}: (i) the $P^2$ algorithm of Jain and Chlamtac~\cite{JainChla1985}, (ii) the random sample based algorithm of Vitter~\cite{Vitt1985}, (iii) the uniform adjustable histogram of Schmeiser and Deutsch~\cite{SchmDeut1977}, and (iv) the data-aligned maximal entropy histogram of Arandjelovi\'c \textit{et al.}~\cite{AranPhamVenk2014,AranPhamVenk2015}. A representative summary of results is shown in Table~\ref{t:resComp}. It can be readily observed that our method and the method of Arandjelovi\'c \textit{et al.} significantly outperformed other approaches. The $P^2$ and equispaced histogram based algorithms performed worst, often producing highly inaccurate estimates. The random sample algorithm of Vitter performed relatively well but still substantially worse than the top two methods. It is interesting to note that the data-aligned maximal entropy histogram of Arandjelovi\'c \textit{et al.} outperformed the proposed method. At first we found this highly surprising given that this algorithm approximates the entire distribution of historical data whereas ours, by design, narrows its focus to the more relevant part of the distribution. We hypothesized that the reason behind this is that the quantile we sought to estimate was insufficiently challenging (not close enough to 1, relative to buffer size). Specifically, our hypothesis stems from the observation that some information is lost by interpolation every time a new datum is added to our buffer. While interpolation is also employed by Arandjelovi\'c \textit{et al.}, when the target quantile is not particularly challenging relative to the buffer size, the number of interpolations performed by the simple data-aligned maximal entropy histogram is lower and its underlying model sufficiently flexible to produce an accurate estimate. Consequently, we hypothesized that the advantages of our method would only be fully exhibited for higher quantiles (needed in applications such as customer wallet estimation~\cite{PerlRossLawrZadr2007}) and we sought to investigate that next.

\begin{table*}
  \centering
  \renewcommand{\arraystretch}{1.55}
  \vspace{7pt}
  \caption{ Comparative experimental results for 0.95-quantile. }
  \vspace{1pt}
  \begin{tabular}{l|c||cccccc}
  \Hline
         &                      & \multicolumn{2}{c}{Stream 1} & \multicolumn{2}{c}{Stream 2} & \multicolumn{2}{c}{Stream 3}\\
  \cline{3-8}
  Method & Bins & Relative      & Absolute         & Relative    & Absolute         & Relative    & Absolute\\
         &                      & $L_1$ error   & $L_\infty$ error & $L_1$ error & $L_\infty$ error & $L_1$ error & $L_\infty$ error\\
  \hline
   Targeted adaptable                                        & 500  &  2.1\% & 1.00e11 & 4.7\% & 24.20 &   5.2\% & 4.8e5\\[-0pt]
   sample (proposed)                                         & 100  &  1.6\% & 1.07e11 & 9.2\% & 54.73 &   3.6\% & 2.89e5\\
   \hline
   Data-aligned max.                                         & 500  &  1.2\% & 3.11e10 & 0.0\% &  2.04 &   0.1\% & 8.11e4\\[-0pt]
   entropy histogram~\cite{AranPhamVenk2014}                 & 100  &  9.6\% & 2.06e11 & 0.0\% &  1.91 &   2.6\% & 3.33e5\\
   \hline
     $P^2$ algorithm~\cite{JainChla1985}                     & n/a  & 15.7\% & 2.77e11 & 3.1\% & 93.04 &  84.2\% & 1.55e6\\
   \hline
    Random sample~\cite{Vitt1985}                            & 500  &  4.6\% & 1.98e11 & 0.7\% & 38.00 &  10.4\% & 5.95e5\\
    \hline
    Equispaced histogram~\cite{SchmDeut1977}                 & 500  & 87.1\% & 1.07e12 & 0.1\% & 80.29 & 675.1\% & 4.39e7 \\
  \Hline
  \end{tabular}
  \label{t:resComp}
\end{table*}

In the second set of experiments we compared our method with the data-aligned maximal entropy histogram of Arandjelovi\'c \textit{et al.} using a series of progressively challenging target quantiles. A summary of the results is shown in Table~\ref{t:resHigh}. It is readily apparent that this set of results fully supports our hypothesis. While our algorithm showed an improvement in performance as the value of the target quantile was increased, the opposite was true for the data-aligned maximal entropy histogram which performed progressively worse. Data set~3 again proved to be the most challenging one, the data-aligned maximal entropy histogram producing grossly inaccurate estimates for quantile values of over $0.99$. For example, on stream~3 for the target quantile of $0.999$ the data-aligned maximal entropy histogram achieved the average relative $L_1$ error of 368.6\%, while the proposed algorithm showed remarkable accuracy and the error of 1.6\%. The same observations can be made by considering the absolute $L_\infty$ error i.e.\ the greatest error in the running quantile estimates, which were respectively 2.35e7 and 3.35e6 -- a difference of approximately an order of magnitude.

\begin{table*}
\vspace{7pt}
  \centering
  \renewcommand{\arraystretch}{1.55}
  \caption{ Comparison of the top two algorithms for high-value quantiles using 100 bins. }
  \vspace{1pt}
  \begin{tabular}{c|c||cc|cc|c}
  \Hline
  \multirow{3}{*}{\rotatebox{90}{Data set~}} & \multirow{3}{*}{Quantile}
                            & \multicolumn{2}{c}{Proposed method} & \multicolumn{2}{|c|}{Data-aligned histogram} &
                            \multirow{3}{*}{\rotatebox{90}{Max}} \multirow{3}{*}{\rotatebox{90}{value to}} \multirow{3}{*}{\rotatebox{90}{quantile}} \multirow{3}{*}{\rotatebox{90}{ratio~}}\\
  \cline{3-6}
  &                            & Relative      & Absolute         & Relative    & Absolute         & \\[-0pt]
                           &   & $L_1$ error   & $L_\infty$ error & $L_1$ error & $L_\infty$ error &\\
  \hline
  \multirow{5}{*}{\rotatebox{90}{Stream 1~}}
                  & 0.9500     & 1.6\%         & 1.07e11          &  9.6\%      & 2.06e11 & 15.8\\
                  & 0.9900     & 1.2\%         & 9.59e10          & 27.9\%      & 5.69e11 & 5.9\\
                  & 0.9950     & 2.1\%         & 9.27e10          & 58.8\%      & 8.48e11 & 4.2\\
                  & 0.9990     & 0.7\%         & 9.80e10          & 48.0\%      & 9.47e11 & 2.1\\
                  & 0.9995     & 0.3\%         & 2.69e10          & 36.8\%      & 8.72e11 & 1.5\\
  \hline
  \multirow{5}{*}{\rotatebox{90}{Stream 2~}}
                  & 0.9500     & 9.2\%         & 54.73            & 0.0\%       & 1.91    & 30.1\\
                  & 0.9900     & 2.4\%         & 26.31            & 0.3\%       & 2.45    &  2.5\\
                  & 0.9950     & 0.3\%         & 6.21             & 0.2\%       & 4.59    &  1.8\\
                  & 0.9990     & 0.2\%         & 16.05            & 0.4\%       & 30.29   &  1.4\\
                  & 0.9995     & 0.2\%         & 20.17            & 2.0\%       & 34.44   &  1.3\\
  \hline
  \multirow{5}{*}{\rotatebox{90}{Stream 3~}}
                  & 0.9500     & 3.6\%         & 2.89e5           & 2.6\%       & 3.33e5  & 520.3\\
                  & 0.9900     & 1.2\%         & 3.32e6           & 2.4\%       & 3.25e5  & 122.7\\
                  & 0.9950     & 1.8\%         & 1.40e6           & 480.5\%     & 1.63e8  &  60.9\\
                  & 0.9990     & 1.6\%         & 3.35e6           & 368.6\%     & 2.35e7  &  11.7\\
                  & 0.9995     & 4.2\%         & 1.30e7           & 364.2\%     & 2.34e8  &   7.2\\
  \Hline
  \end{tabular}
  \label{t:resHigh}
\end{table*}

\begin{figure}[htb]
  \vspace{3pt}
  \centering
  \includegraphics[width=0.49\textwidth]{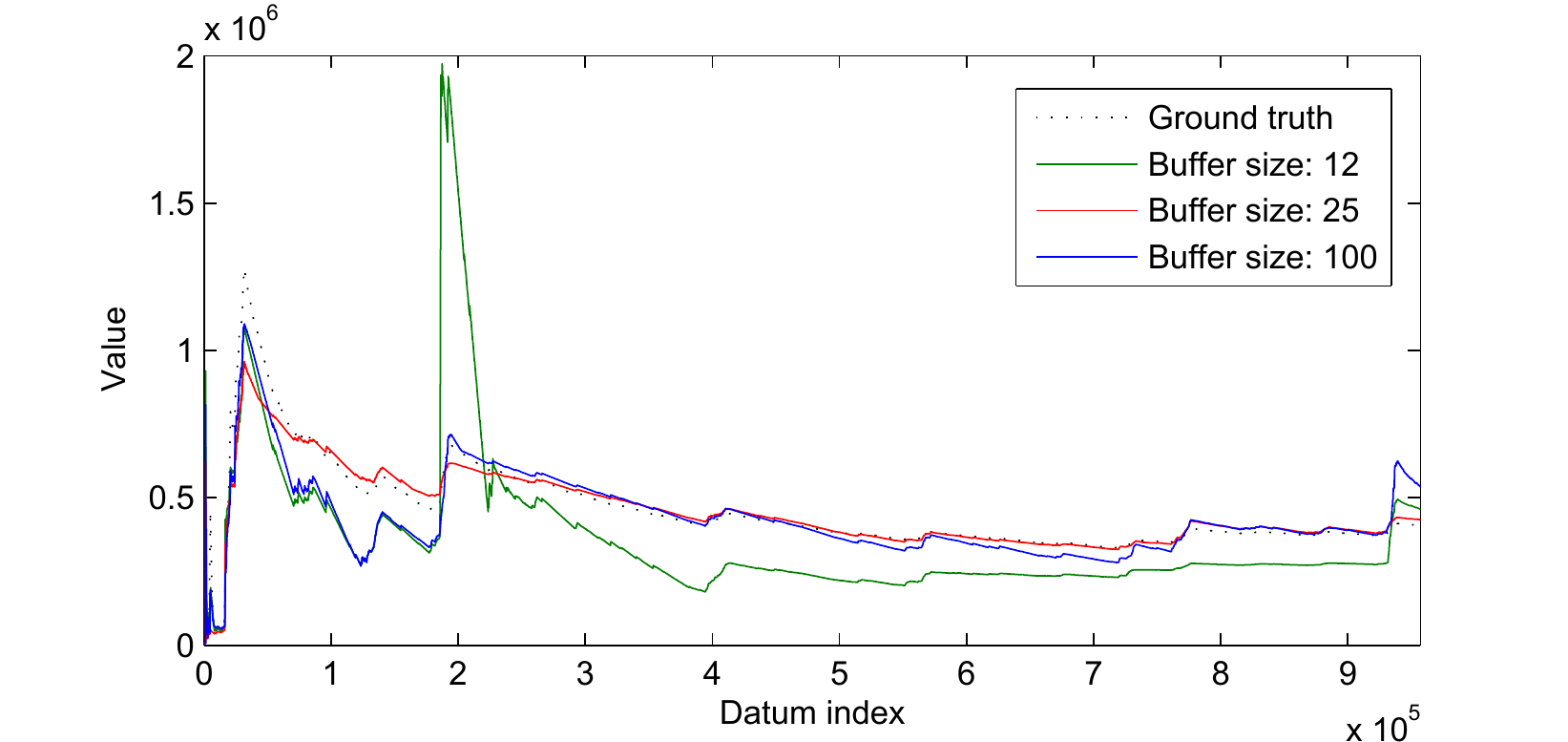}  
  \caption{ An example running ground truth of the target quantile ($q=0.95$) and the estimates of our algorithm for different bin sizes on data stream~3. It is remarkable to observe that our method achieved a consistently highly accurate estimate even when the available buffer capacity was severely restricted (down to only 12 bins).}
  \label{f:resX3095}
\end{figure}

\begin{figure}[htb]
  \centering
  \vspace{3pt}
  \includegraphics[width=0.49\textwidth]{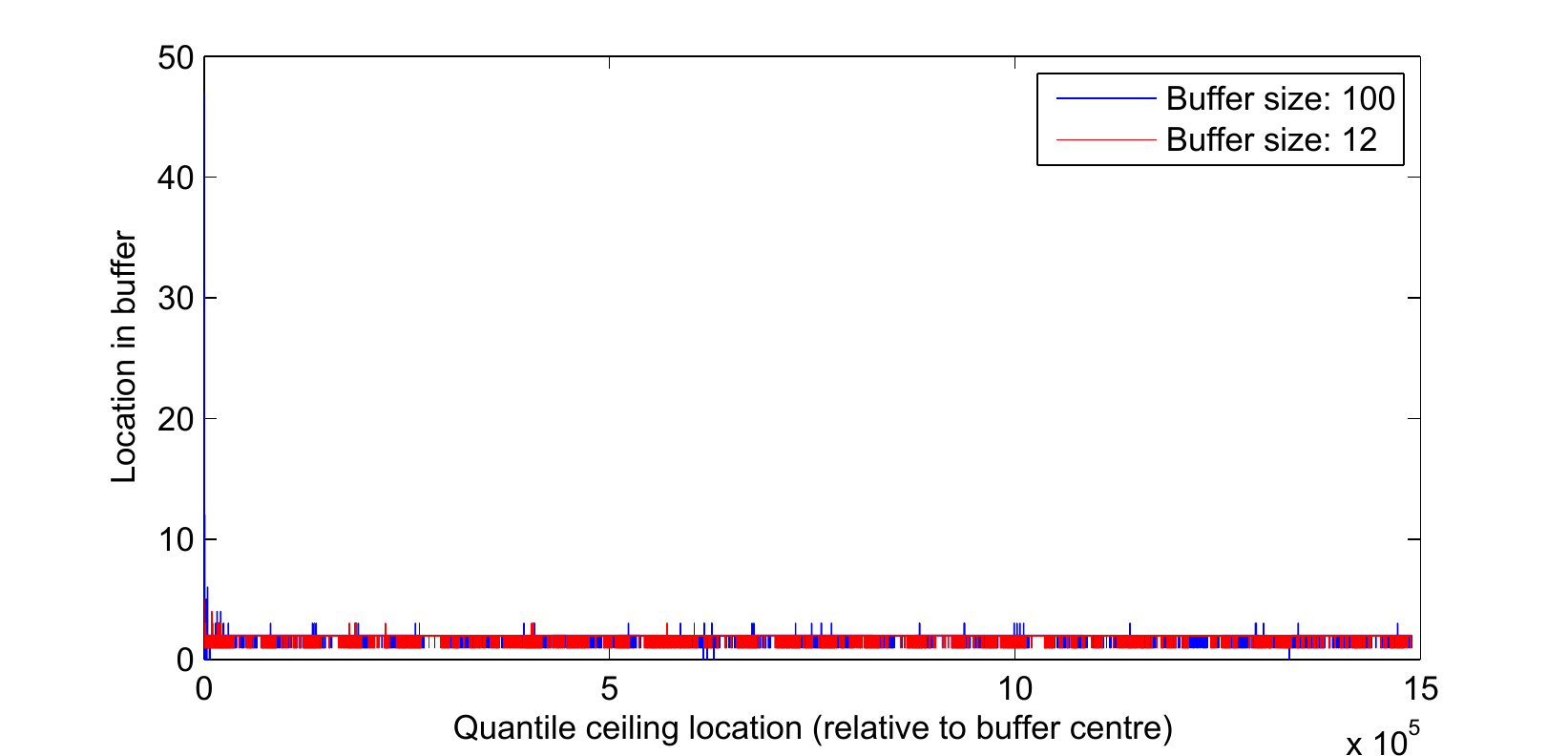}  
  \caption{ Our algorithm is highly successful in achieving one of the key ideas behind the method, that of adapting the data sample retained in the buffer so as to maintain the position of the current quantile estimate in the buffer as close to its centre as possible (see Section~\ref{sss:constraints}). Both in the case of a buffer with the capacity of 100 and 12 (the results for only two buffer sizes are shown to reduce clutter), the central positioning of the quantile estimate is maintained very tightly throughout the processing of the stream.}
  \label{f:resX3centres}
\end{figure}

\begin{figure}[htb]
  \centering
  \vspace{3pt}
  \includegraphics[width=0.49\textwidth]{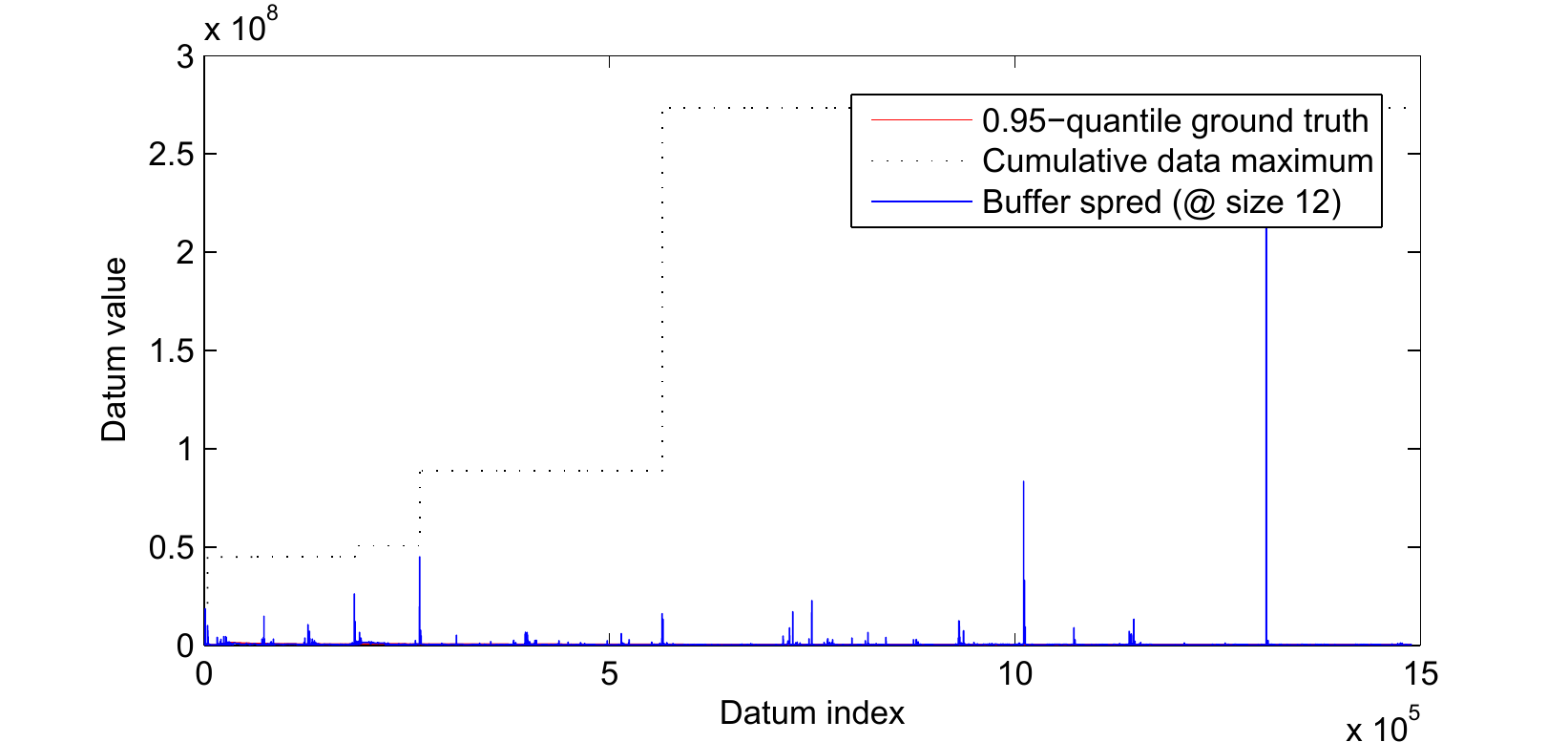}
  \caption{ The success of our algorithm in achieving tight sampling of the data distribution around the target quantile -- unlike the random sample based algorithm of Vitter~\cite{Vitt1985} or the uniform adjustable histogram of Schmeiser and Deutsch~\cite{SchmDeut1977} which retain a sample from a wide range of values, our method utilizes the available memory efficiently by focusing on a narrow spread of values around the current quantile estimate. While the spread of values in the buffer experiences intermittent and transient increases when there is a burst of high valued data points in preparation for a potentially large quantile change, thereafter it quickly adapts to the correct part of the distribution.}
  \label{f:resX3spread}
\end{figure}

Table~\ref{t:resHigh} also includes a column (right-most) showing the ratio of the maximal stream value and the ground truth for the target quantile. We sought to examine if a particularly high ratio predicts poor performance of the data-aligned maximal entropy histogram, which may be expected given that throughout its operation the algorithm approximates the entire distribution of historical data. We found this not to be the case which can be explained by the allocation of bin ranges according to the maximum entropy principle and the alignment of the bin boundaries with data; please see the original publication for a detailed description of the method~\cite{AranPhamVenk2014}.

\begin{figure}[htb]
  \centering
  \vspace{10pt}
  \includegraphics[width=0.49\textwidth]{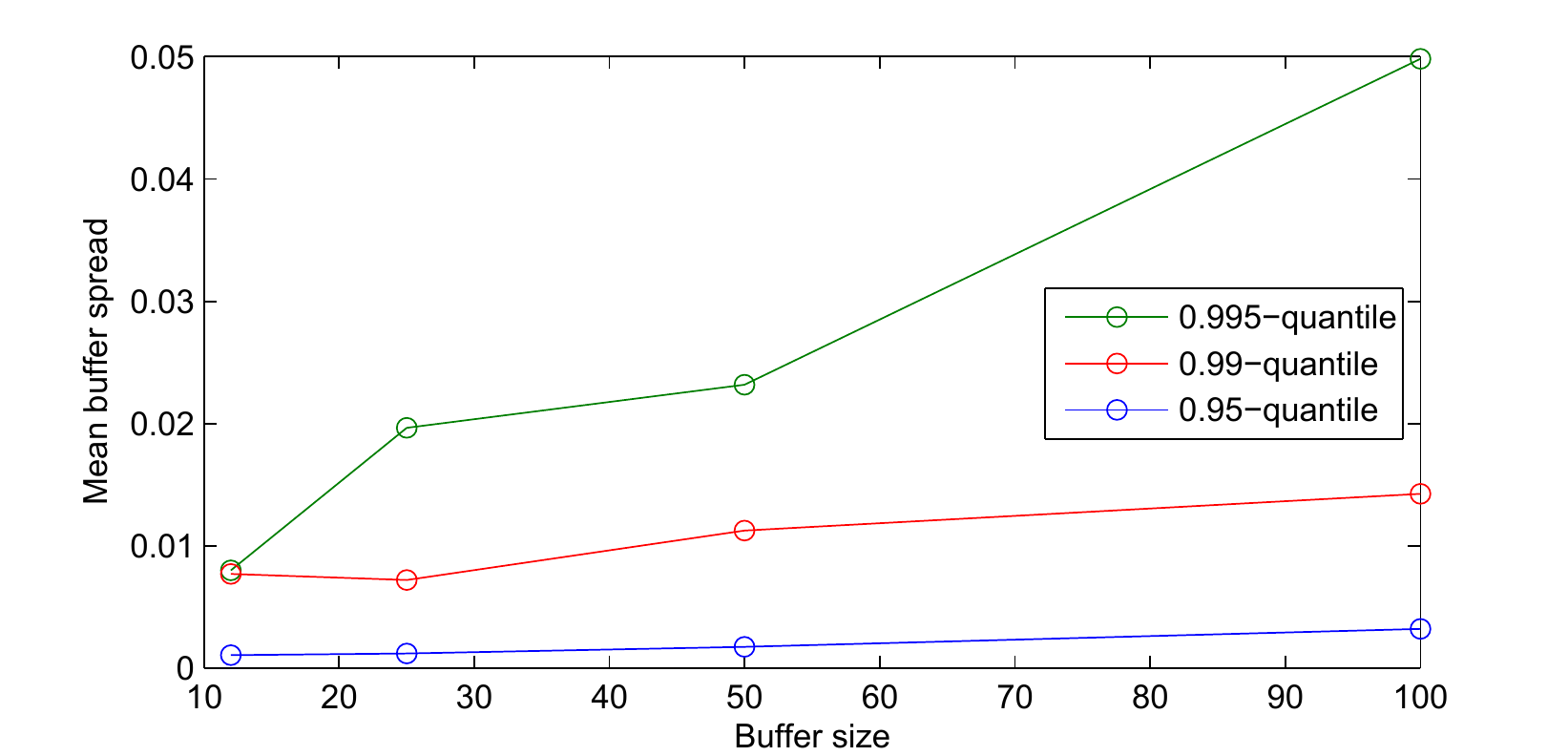}
  \vspace{3pt}
  \caption{ The variation in the mean buffer spread with the buffer size and target quantile on data stream~3.}
  \label{f:resX3spreadMean}
  \vspace{0pt}
\end{figure}

\begin{figure}[htb]
  \centering
  \includegraphics[width=0.49\textwidth]{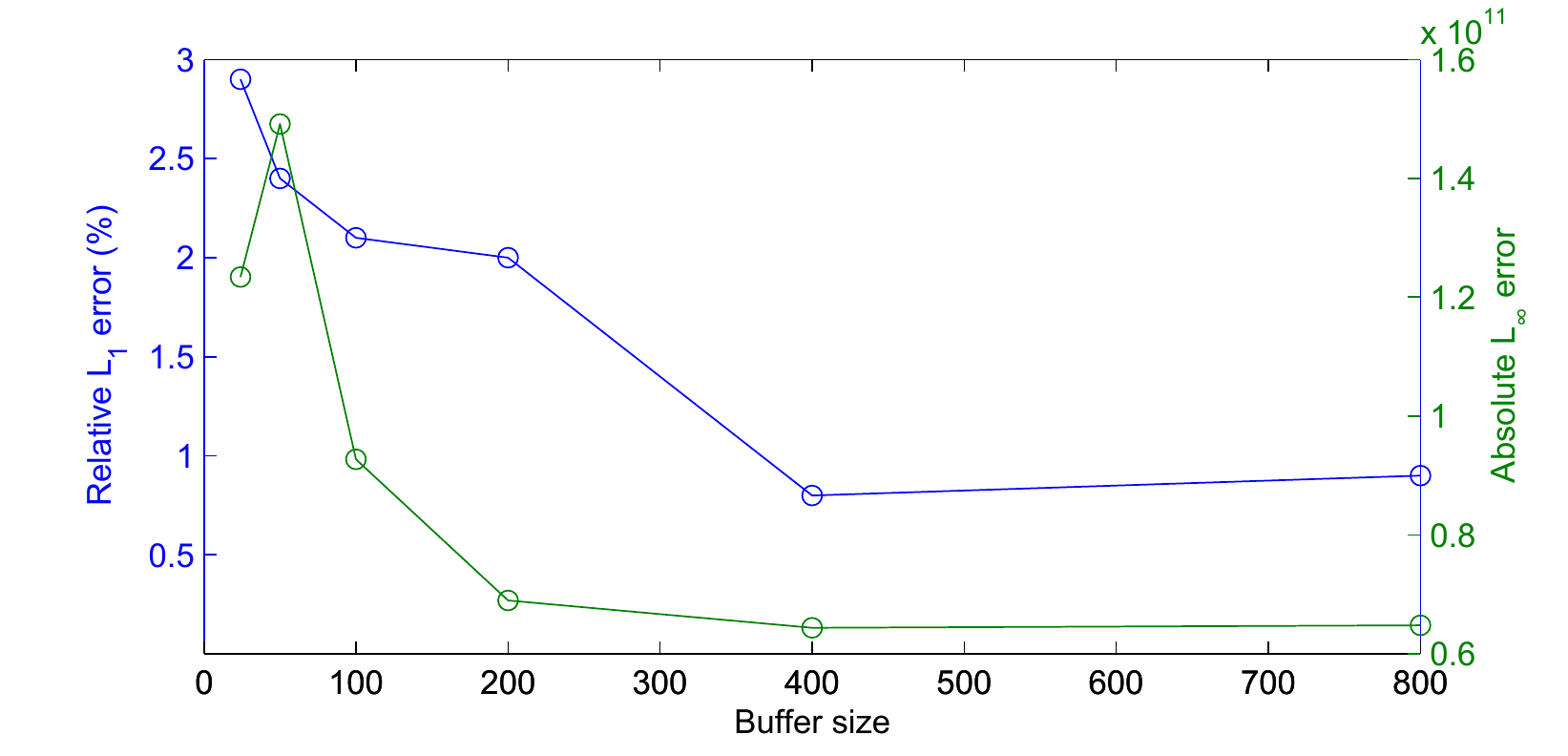}
  \vspace{3pt}
  \caption{ The variation in the accuracy of our algorithm's estimate with the buffer size on data stream~3. Unlike any of the existing algorithms, our method exhibits very gradual and graceful degradation in performance, and still achieves remarkable accuracy even with a severely restricted buffer capacity.}
  \label{f:errSize}
  \vspace{10pt}
\end{figure}

Lastly, we sought to analyse the performance of the proposed method in additional detail. Figure~\ref{f:resX3095} shows on an example the running ground truth of the target quantile ($q=0.95$) and the estimates of our algorithm for different bin sizes on the most challenging data stream~3. It is remarkable to observe that our method consistently achieved a highly accurate estimate even when the available buffer capacity was severely restricted (to 12 bins). In Figure~\ref{f:resX3centres} the same example run was used to illustrate the success of our algorithm in achieving one of the key ideas behind the method, that of adapting the data sample retained in the buffer so as to maintain the position of the current quantile estimate in the buffer as close to its centre as possible (see Section~\ref{sss:constraints}). As the plot clearly shows, both in the case of a buffer with the capacity of 100 and 12 (the results for only two buffer sizes are shown to reduce clutter), the central positioning of the quantile estimate is maintained very tightly throughout the processing of the stream. Similarly, the success of our algorithm in achieving tight sampling of the data distribution around the target quantile is illustrated in the plot in Figure~\ref{f:resX3spread}. This plot shows that unlike the random sample based algorithm of Vitter~\cite{Vitt1985} or the uniform adjustable histogram of Schmeiser and Deutsch~\cite{SchmDeut1977} which retain a sample from a wide range of values, our method utilizes the available memory efficiently by focusing on a narrow spread of values around the current quantile estimate. Note that the spread of values in the buffer experiences intermittent and transient increases when there is a burst of high valued data points in preparation for a potentially large quantile change, but thereafter quickly adapts to the correct part of the distribution. The variation in the mean buffer spread with the buffer size and target quantile is shown in Figure~\ref{f:resX3spreadMean}. Lastly, the variation in the accuracy of our algorithm's estimate with the buffer size is analysed in Figure~\ref{f:errSize}. Unlike any of the existing algorithms, our method exhibits very gradual and graceful degradation in performance, and still achieves remarkable accuracy even with a severely restricted buffer capacity.

\section{Summary and conclusions}
In this paper we described a novel algorithm for the estimation of a quantile of a data stream when the available working memory is limited (constant), prohibiting the storage of all historical data. This problem is ubiquitous in computer vision and signal processing, and has been addressed by a number of researchers in the past. We showed that a major shortcoming of the existing methods lies in their usually implicit assumption that the data is being generated by a stationary process. This assumption is invalidated in most practical applications, as we illustrated using real-world data.

Evaluated on three large data streams extracted from CCTV footage, our algorithm was vastly superior in comparison with the existing alternatives. The highly non-stationary nature of the data was shown to cause major problems to previous methods, often leading to grossly inaccurate quantile estimates; in contrast, our method was virtually unaffected by it. What is more, our experiments demonstrate that the superior performance of our algorithm can be maintained effectively while drastically reducing the working memory size in comparison with the methods from the literature.

{\linespread{1.25}
  \bibliographystyle{ieee}
  \bibliography{../../../my_bibliography}
}
\balance

\end{document}